\documentclass[10pt,twocolumn,letterpaper]{article}

\usepackage{3dv}
\usepackage{times}
\usepackage{epsfig}
\usepackage{graphicx}
\usepackage{amsmath}
\usepackage{amssymb}

\usepackage{soul}
\usepackage{subcaption}
\usepackage{xcolor}

\usepackage[pagebackref=true,breaklinks=true,letterpaper=true,colorlinks,bookmarks=false]{hyperref}

\threedvfinalcopy 

\def\threedvPaperID{389} 

\ifthreedvfinal\fi
\begin{document}

\title{MaskNet: A Fully-Convolutional Network to Estimate Inlier Points}

\author{\fontsize{10}{15}\textbf{Vinit Sarode}\thanks{equal contribution\newline \text{Correspondence to \{vinitsarode5, animeshdhagat\}@gmail.com}},
\textbf{Animesh Dhagat}\textsuperscript{*}, 
\textbf{Rangaprasad Arun Srivatsan},
\fontsize{10}{15}\textbf{Nicolas Zevallos},
\textbf{Simon Lucey},
\textbf{Howie Choset}\\
Carnegie Mellon University\\
{\tt\small vinitsarode5@gmail.com,}
{\tt\small adhagat@andrew.cmu.edu}
}

\maketitle

\begin{abstract}
Point clouds have grown in importance in the way computers perceive the world. From LIDAR sensors in autonomous cars and drones to the time of flight and stereo vision systems in our phones, point clouds are everywhere. Despite their ubiquity, point clouds in the real world are often missing points because of sensor limitations or occlusions, or contain extraneous points from sensor noise or artifacts. These problems challenge algorithms that require computing correspondences between a pair of point clouds. Therefore, this paper presents a fully-convolutional neural network that identifies which points in one point cloud are most similar (inliers) to the points in another. We show improvements in learning-based and classical point cloud registration approaches when retrofitted with our network. We demonstrate these improvements on synthetic and real-world datasets. Finally, our network produces impressive results on test datasets that were unseen during training, thus exhibiting generalizability. Code and videos are available at~\url{https://github.com/vinits5/masknet}
\end{abstract}

\section{Introduction}
Point clouds are unordered sets of points in 3D space, usually describing the surface of an object or a scene.
Recently they have been used to recognize~\cite{PointNet}, locate~\cite{Chen2017Multiview3O} and track objects~\cite{Bertinetto2016FullyConvolutionalSN} in a scene or be stitched together to form more complete point clouds~\cite{Chu2011MultiviewPC}. When being used in cluttered spaces, point clouds describe only parts of the object/scene that are visible to the sensor and not covered by occlusions. In addition, sensor noise, reflective surfaces, or other artifacts can sometimes produce points in the point cloud which do not correspond to any surface on the object or in the scene. Point clouds with missing data as well as those having extraneous points pose challenges to point cloud processing algorithms such as registration~\cite{INTRO:ICP} \cite{yaoki2019pointnetlk} and tracking~\cite{Bertinetto2016FullyConvolutionalSN}. As a result, it is important to identify which points need to be considered `inliers' and which points need to be discarded and deemed as `outliers'.

\begin{figure}[t!]
    \centering
    \includegraphics[width=\columnwidth]{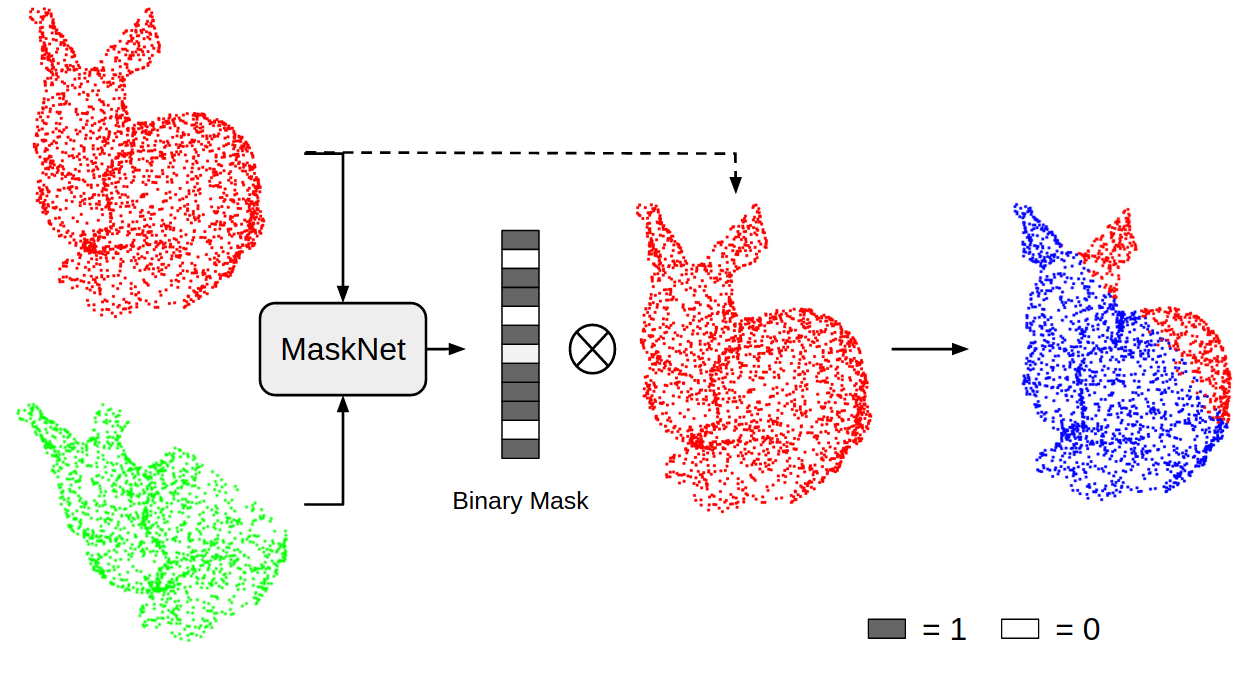}
    \caption{
    MaskNet estimating inliers (shown on the right in blue) for a pair of point clouds (shown on the left). MaskNet finds a Boolean vector \textit{mask} that only retains inlier points from point cloud in red which most closely approximate the shape of the point cloud in green.
    }
    \label{fig:firstfig}
\end{figure}

Prior works on inlier/outlier detection have considered taking random consensus (RANSAC)~\cite{amo_fpcs_sig_08}, finding geometric primitives such as planes and cylinders~\cite{li2017improved,Tombari2010UniqueSO}, finding most probable correspondences~\cite{billings2015iterative,bustos2017guaranteed}, or using globally optimal alignment techniques~\cite{DBLP:journals/corr/YangLCJ16,izatt2020globally,srivatsan2019globally}. However, most of these works do not scale well with the number of points in the point clouds, and therefore are generally computationally expensive for most real-time applications. 

In this work, we leverage recent advancements in deep learning-based point cloud representations~\cite{PointNet}, to perform learning-based inlier estimation.
Given a template point cloud and a source point cloud (shown by red and green points in Fig.~\ref{fig:firstfig} respectively), we create a network that is trained to identify which points from the template are inliers, so that these points closely describe the same part of the object/scene geometry as the source point cloud~(the blue points in Fig.~\ref{fig:firstfig} are the inliers).
In other words, the network learns to `mask-out' outliers from the template point cloud, hence we call our approach MaskNet.

We evaluate MaskNet on synthetic~\cite{wu20153d} as well as real-world datasets~\cite{zeng20163dmatch,armeni_cvpr16} and compare with state-of-the-art approaches. Further, we demonstrate the benefit of using MaskNet as a preprocessing step for point cloud registration algorithms. Removing outliers in particular improves the registration accuracy of popular deep learning-based approaches such as PointNetLK~\cite{yaoki2019pointnetlk}, and deep closest point (DCP)~\cite{Wang_2019_ICCV}. Finally, MaskNet shows remarkable generalization within and across datasets without the need for additional fine tuning. 

In this paper, we review prior work and define the problem as - finding inlier points in a given pair of point clouds describing the same object, where one of them (source) has missing points compared to the other (template) - in Sec.~\ref{related_work}. In Sec.~\ref{masknet} we describe our approach for finding inlier points and in Sec.~\ref{experiment} we discuss the veracity of our approach through experiments on synthetic and real-world datasets.

\newline
\section{Related Work}
\label{related_work}
\paragraph{Classical Inlier/Outlier Estimation Approaches}  Given a pair of point clouds, inlier estimation determines which points in one point cloud are most similar to the points in another. Estimating inlier point-pairs is a well documented problem that finds application in registration~\cite{choy2020deep}, object detection~\cite{Sveier2017ObjectDI}, and flow estimation~\cite{Liu2019FlowNet3DLS}. Locally optimal methods achieve this by finding all point-pairs between point clouds, and retaining only the inliers by using algorithms such as RANSAC~\cite{schnabel2007efficient,pankaj2016robust}. They iteratively sample random point-pairs until the inlier point-pairs minimize the misalignment between the point clouds. However, these methods are very time consuming, especially when the initial misalignment between the point clouds is very high. RANSAC-based approaches also exhibit slow convergence and low accuracy with large number of outliers.

Other works have posed inlier estimation as an optimization problem and used branch-and-bound techniques~\cite{Yang2013GoICPS3}, Gaussian mixture models~\cite{myronenko2010point}, mixed integer programming (MIP)~\cite{Izatt2017GloballyOO}, semi-definite programming~\cite{chaudhury2015global} and maximal clique selection~\cite{bustos2019practical}. These methods provide guarantees on robust detection of all the inliers. With the exception of TEASER++~\cite{yang2020teaser} and ~\cite{Dym2019LinearlyCQ}, most of these conventional optimization-based approaches are computationally expensive and cannot be deployed for any practical realtime applications.

A fast-growing vein in inlier estimation has been through finding point-based features, and using them to significantly reduce the number of outliers. Local geometry-based features are computed from a combination of 3D point coordinates and surface normals~\cite{bustos2017guaranteed,Rusu2009FastPF}. Similarly other works have introduced hand-crafted features~\cite{Learningrusu2008}. Compared to globally optimal approaches, feature-based inlier detection approaches are computationally faster, but more prone to failure as they are sensitive to the specific choice of geometric features that are chosen.

\paragraph{Learning in Point Clouds}
Inspired by classical feature-based inlier estimation techniques, Qi et al.~\cite{PointNet} introduced PointNet, a learning-based approach to learning task specific features from point clouds. PointNet~\cite{PointNet} paved the way for using unordered point clouds in a learning paradigm. Recent works such as PointNet++ ~\cite{qi2017pointnetplusplus}, DGCNN~\cite{wang2019dynamic},Deep sets~\cite{zaheer2017deep}, PointCNN~\cite{li2018pointcnn}, point pillars~\cite{lang2019pointpillars} and PCPNet~\cite{guerrero2018pcpnet} improve the performance of PointNet by generating features that consider local neighbourhoods of points. More recently, a point cleaning network~\cite{rakotosaona2020pointcleannet} was developed based on the PCPNet. PointCleanNet  estimates robust local features and use this information to denoise the point cloud. Another variant of learning-based inlier estimation includes SampleNet~\cite{lang2019samplenet,Dovrat_2019} which finds only a small subset of inliers. 

In addition to supervised learning techniques such as PointNet, other self-supervised~\cite{DBLP:journals/corr/abs-1901-08396}, and unsupervised~\cite{8885536} feature-learning techniques have been introduced. While all these algorithms continue to remain locally optimal, they are faster than conventional geometry-based methods, owing to computationally inexpensive matrix operations. However, as the point clouds become denser, the neural networks suffer as much as their hand-crafted counterparts in terms of computational complexity.

\paragraph{Point Cloud Registration} 
An important application in computer vision that is impacted heavily by the presence of outliers in the point cloud data is that of point cloud registration. Outliers adversely affect both classical~\cite{INTRO:ICP,rusinkiewicz2001efficient,Mellado2014Super4F,amo_fpcs_sig_08} and deep learning-based registration methods~\cite{yaoki2019pointnetlk,vsarode2019pcrnet,Wang_2019_ICCV}.
Most of the popular deep learning registration-based methods such as PointNetLK~\cite{yaoki2019pointnetlk}, PCRNet~\cite{vsarode2019pcrnet} and DCP~\cite{Wang_2019_ICCV} work on the assumption that all points in the point clouds are inliers by default. Naturally, they perform poorly when one of the point clouds has missing points, as in the case of partial point clouds. Deep learning-based methods developed specifically to handle partial point cloud registration such as PRNet~\cite{Wang_2019_NeurIPS}, and RPM-Net~\cite{yew2020-RPMNet} have deeper network architectures that learn to handle partial point clouds by explicitly estimating point-correspondences. Unfortunately, these methods do not scale well with the increasing number of points, as the size of the correspondence parameters predicted by the network grows polynomially. They also rely heavily on being explicitly trained with partial point cloud data. Other methods such as deep global registration (DGR)~\cite{choy2020deep} and Multi-view registration~\cite{Gojcic2020LearningM3} use neural networks to filter outliers from a given set of possible correspondences. Such methods are limited to only estimating a subset of inlier points.

\section{MaskNet}
\label{masknet}
\begin{figure*}[t!]
    \centering
    \includegraphics[width=0.9\linewidth]{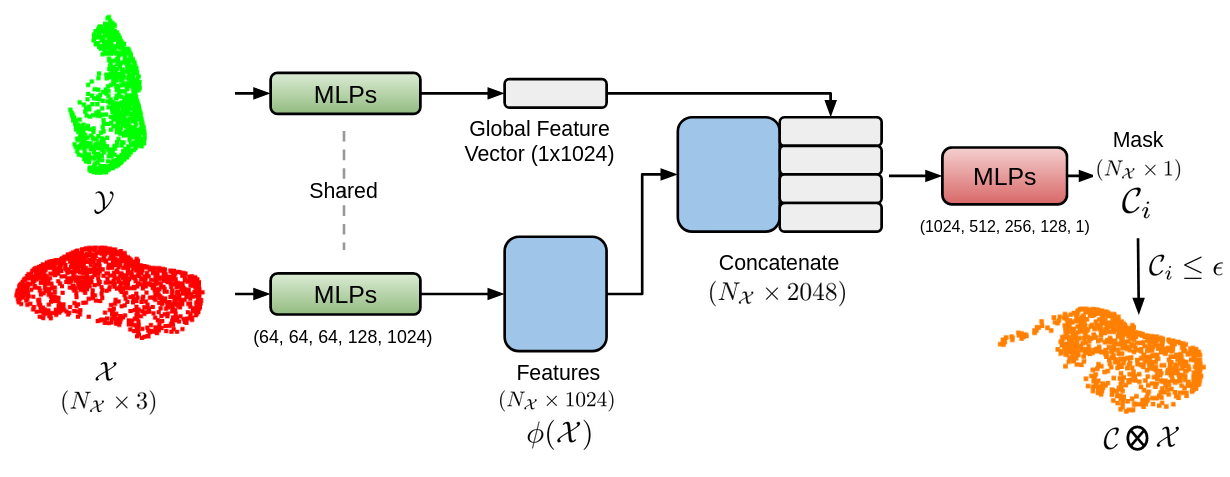}
    \caption{MaskNet Architecture: The model consists of five MLPs of size (64, 64, 64, 128, 1024). The source ($\mathcal{Y}$) and template ($\mathcal{X}$) point clouds are sent as input through a twin set of MLPs, arranged in a Siamese architecture. Using a max-pooling function, we obtain global features for the source point cloud. Weights are shared between MLPs. These features are concatenated and provided as an input to five fully convolutional layers of size (1024, 512, 256, 128), and an output layer of size $1$. A sigmoid function is applied on the output to produce the mask ($\mathcal{C}$ ).
}
    \label{fig:MaskNet_architecture}
\end{figure*}
In this section we discuss the mathematical formulation of MaskNet and describe how MaskNet can be used to improve the task of point cloud registration.

\subsection{Mathematical Formulation}
\label{sec:mathematical_foundation}

Let us consider two point clouds $\mathcal{X}$ and $\mathcal{Y}$, where $\mathcal{X} = [\boldsymbol{x}_1, \boldsymbol{x}_2, ..., \boldsymbol{x}_{N_{\mathcal{X}}}] \in \mathbb{R}^{3 \times N_{\mathcal{X}}}$, $\mathcal{Y} = [\boldsymbol{y}_1, \boldsymbol{y}_2, ..., \boldsymbol{y}_{N_y}] \in \mathbb{R}^{3 \times N_y}$, and $N_{\mathcal{X}} \geq N_y$. 
Let us consider a binary vector $\mathcal{C} \in \mathbb{R}^{N_{x} \times 1}$ ($\mathcal{C}_i \in \{0, 1\} \quad \forall i \in [1, ..., N_{\mathcal{X}}]$) and an operator $\otimes$, defined as follows:
\begin{equation}
\mathcal{X}_1 =\mathcal{C} \otimes\mathcal{X},
\end{equation}
where $\mathcal{X}_1\subseteq\mathcal{X}$ and $\boldsymbol{x}_i \in \mathcal{X}_1$ if $\mathcal{C}_i=1$. In other words, the set $\mathcal{X}_1$ contains a subset of points from the set $\mathcal{X}$ corresponding to $\mathcal{C}_i=1$. Effectively the binary vector $\mathcal{C}$ acts as a mask and removes points from the set $\mathcal{X}$. We aim to mask the points from $\mathcal{X}$ such that the resulting point cloud $\mathcal{X}_1$ represents the same overall geometry as $\mathcal{Y}$. We use the K-dimensional vector encoding of PointNet~\cite{PointNet}, $\phi : \mathbb{R}^{3\times N} \rightarrow \mathbb{R}^{K}$ as the metric for comparing $\mathcal{X}_1$ and $\mathcal{Y}$. From PointNetLK~\cite{yaoki2019pointnetlk} we know that the PointNet encoding can be sensitive to large misalignment between the point clouds and hence the following condition holds:
\begin{equation}
    \phi(\mathcal{C} \otimes \mathcal{X}) = \phi(\mathbf{R}\mathcal{Y} + \mathbf{t}),
    \label{eq:feature_compare}
\end{equation}
where $\mathbf{R} \in SO(3)$ is the rotation and $\mathbf{t} \in \mathbb{R}^3$ is the translation between $\mathcal{X}$ and $\mathcal{Y}$. However, since $\mathbf{R}$ and $\mathbf{t}$ are unknown, we use the following weaker condition to relate $\mathcal{X}_1$ and $\mathcal{Y}$:
\begin{equation}
    \phi(\mathcal{C} \otimes \mathcal{X}) \approx \phi(\mathcal{Y}).
    \label{eq:feature_compare_approx}
\end{equation}
We show later in Fig.~\ref{fig:iteration_results}, that it is possible to iteratively improve the estimate of $\mathcal{C}$. We substitute $\mathcal{C}$ from Eq.~\ref{eq:feature_compare_approx} into Eq.~\ref{eq:feature_compare} and find $\mathbf{R}$ and $\mathbf{t}$ as described in Sec.~\ref{sc:registration}. We then replace $\mathcal{Y}$ in Eq.~\ref{eq:feature_compare_approx} with $\mathbf{R}\mathcal{Y} + \mathbf{t}$ and repeat this process until convergence. 
We learn $\mathcal{C}$ using MaskNet, $f(\cdot)$, as follows:
\begin{equation}
    \mathcal{C} = f(\phi(\mathcal{X}), \phi(\mathcal{Y})).   
    \label{eq:masknet_function}
\end{equation}

Discrete outputs from a neural network during training induce discontinuities during back propagation and prevents MaskNet from producing a binary vector as output. Instead, MaskNet predicts a vector $\mathcal{C}^*$ such that $\mathcal{C}^*_i \in [0, 1] \quad \forall i \in [1, ..., N_{\mathcal{X}}]$. During evaluation, a binary vector is computed by applying a threshold ($\epsilon$) such that,
\begin{equation}
    \mathcal{C} = \left\{
    \begin{array}{@{}ll@{}}
        1, \quad  \text{ if } \quad \mathcal{C}^*_i \geq \epsilon
    \end{array}\right\}
    \quad \forall i \in [1 , N_{\mathcal{X}}]
\end{equation}

\subsection{Point Feature Encoding}
\label{sec:point cloud encoder}
PointNet uses a set of multi-layer perceptrons (MLPs) to encode each 3D point in a higher dimensional feature vector. Similar to PointNet, MaskNet (see Fig.~\ref{fig:MaskNet_architecture}) uses a set of MLPs of size (64, 64, 64, 128, 1024) to estimate the feature vectors ($\phi(\mathcal{X})$, $\phi(\mathcal{Y})$) of input point clouds. Prior works such as PCN~\cite{yuan2018pcn} and FoldingNet~\cite{yang2018foldingnet} have shown the efficacy of PointNet feature vectors for point cloud completion by comparing the features of partial input point cloud with the features of 2D point grids to create local patches of the complete output point cloud. Taking inspiration from these methods, MaskNet estimates the inliers by concatenating the feature vectors of $\mathcal{X}$ and $\mathcal{Y}$ as follows,
\begin{equation}
    \mathcal{C}^* = sigmoid\left(h\left(
    \begin{bmatrix}
     & g(\phi(\mathcal{Y}))\\
    \phi(\mathcal{X}) & \vdots\\
    & g(\phi(\mathcal{Y}))\\
    \end{bmatrix}_{N_{\mathcal{X}}\times 2048}\right)\right)
\end{equation}
where $h(\cdot)$ represents a set of MLPs of size (1024, 512, 256, 128, 1) used to create the mask $(\mathcal{C})$, $g(\cdot)$ represents the symmetry function (max-pooling operation) where $g(\phi(\mathcal{Y}))\in \mathbb{R}^{1\times1024}$, and $\phi(\mathcal{X})\in \mathbb{R}^{N_{\mathcal{X}}\times1024}$. In the last layer of $h(\cdot)$, we use a sigmoid activation function to enforce $\mathcal{C}^*_i \in [0, 1] \quad \forall i \in [1, ..., N_\mathcal{X}]$.

\subsection{Loss Function}
\label{sec:loss_function}
During training, the loss function of  MaskNet measures the difference between the predicted mask and the ground truth mask, defined as a mean squared error:
\begin{equation}
    Loss = \sum_{i=1}^{N_{\mathcal{X}}}||{\mathbf{C_i} - \mathbf{C_i}^{gt}}||_2,
\end{equation}
where $\textbf{C}$ represents the mask predicted by network and $\textbf{C}^{gt}$  represents the ground truth mask.

\subsection{MaskNet for Registration}
\label{sc:registration}
\begin{figure}[htbp]
    \begin{center}
        \includegraphics[width=\linewidth]{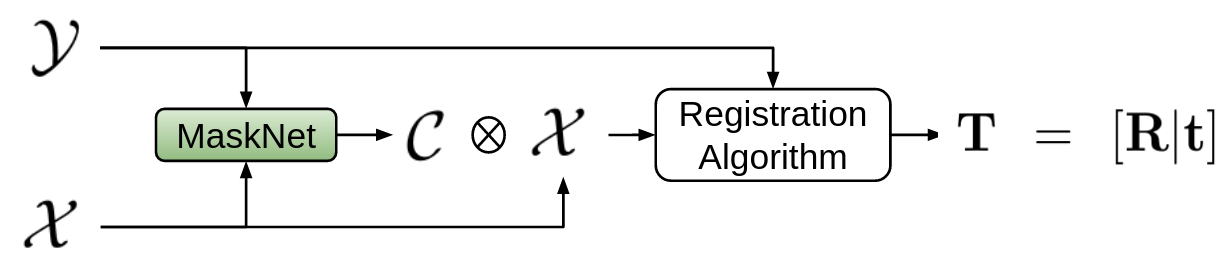}
    \end{center}
   \caption{Registration pipeline: A template point cloud, $\mathcal{X}$ and a source point cloud, $\mathcal{Y}$, are provided to a pre-trained MaskNet. The mask, $\mathcal{C}$ obtained from the network is applied to the template to obtain a partial point cloud, \mbox{$\mathcal{X}_1=\mathcal{C}\otimes\mathcal{X}$}, which is the set of inlier points. $\mathcal{X}_1$ and $\mathcal{Y}$ are provided to a registration algorithm to produce $\mathbf{R}$ \& $\mathbf{t}$. In the case of learning-based registration algorithms, a pre-trained network can be used.}
\label{fig:masknet_registration_framework}
\end{figure}

MaskNet can be used as an add-on module with any registration algorithm to estimate rotation $\mathbf{R}$ and translation $\mathbf{t}$ between a pair of point clouds (see Fig.~\ref{fig:masknet_registration_framework}). For instance, upon finding the mask $\mathcal{C}$ using MaskNet, we can substitute it into Eq.~\ref{eq:feature_compare} and iteratively estimate $\mathbf{R}$ and $\mathbf{t}$ using the Lucas-Kanade algorithm as in the case of PointNetLK~\cite{yaoki2019pointnetlk}. Alternatively, one could apply $\mathcal{C}$ to $\mathcal{X}$ to obtain $\mathcal{X}_1=\mathcal{C}\otimes\mathcal{X}$. Following this, $\mathcal{X}_1$ and $\mathcal{Y}$ can be registered using PCRNet~\cite{vsarode2019pcrnet}, DCP~\cite{Wang_2019_ICCV}, ICP~\cite{INTRO:ICP}, RPM-Net~\cite{yew2020-RPMNet}, and PRNet~\cite{Wang_2019_NeurIPS}.

\subsection{MaskNet for Denoising}
MaskNet can also be used to detect noise and remove outliers from a given point cloud $\mathcal{Y}$, using a noise-free template point cloud of the same category, $\mathcal{X}$, by a simple reformulation. Rather than removing points from $\mathcal{X}$, $\mathcal{C}$ can be applied as a mask to remove outliers present in the set $\mathcal{Y}$. The resulting point cloud $\mathcal{Y}_1 = \mathcal{C} \otimes \mathcal{Y}$ represents the same overall geometry as $\mathcal{X}$. In other words, we flip $\mathcal{X}$ and $\mathcal{Y}$ in Eq. \ref{eq:feature_compare_approx} as follows,
\begin{equation}
    \phi(\mathcal{X}) \approx \phi(\mathcal{C} \otimes \mathcal{Y}).
\end{equation}
Similar to Eq. \ref{eq:masknet_function}, we learn $\mathcal{C}$ using MaskNet.

\section{Experiments}
\label{experiment}
To validate the ability of MaskNet to predict inlier points, Sec. \ref{sec:mask_evaluation} discusses the use of precision as a metric for comparing predicted inlier points to the ground truth inlier points. We then use this precision metric to compare MaskNet with other methods such as RPM-Net and PRNet which explicitly find inlier points based on correspondences. In Sec. \ref{sec:denoising} and Sec.~\ref{sec:registration} we demonstrate the effectiveness of using inlier estimation for the tasks of point cloud denoising and registration of partial point clouds, respectively, on a synthetic dataset ~\cite{wu20153d}. The generalizability of MaskNet is demonstrated on synthetic as well as real-world datasets (S3DIS~\cite{armeni_cvpr16} and 3DMatch~\cite{zeng20163dmatch}) in Sec.~\ref{sec:generalization} and Sec.~\ref{sec:real_world_experiments}.

\subsection{Inlier detection}
\label{sec:mask_evaluation}
In this section we use the ModelNet40 dataset~\cite{wu20153d} consisting of 9843 meshed CAD models of 40 different object categories for training and 2486 models for evaluation. We follow the protocol used in~\cite{PointNet} to create a set of point clouds where 1024 points are uniformly sampled from each mesh model and are scaled to fit in a unit sphere. A random transformation is applied to each of these point clouds with a rotation in $[0, 45]^\circ$ and translation in [-1, 1] units along each axis to create pairs of point clouds. Partial scans of point clouds are obtained by simulating a physical sensor model at a chosen view point. We do this by selecting a random view point and then choosing a set of points from the surface of the object facing the sensor. To obtain these points, we compute nearest neighbors on the surface of object from the view point.   
A ground truth Boolean vector is computed using the indices of nearest neighbours having 1 for selected neighbours and 0 for other points. During training, this Boolean vector is used as a supervision on the predicted mask.

MaskNet is trained for 300 epochs using a learning rate of $10^{-4}$ and batch size $32$. The network parameters are updated with Adam Optimizer on a single NVIDIA GeForce GTX 1070 GPU and an Intel Core i7 CPU at 4.0GHz. We follow the same settings of training for all the experiments.

\begin{figure}
    \begin{center}
        \includegraphics[width=\linewidth]{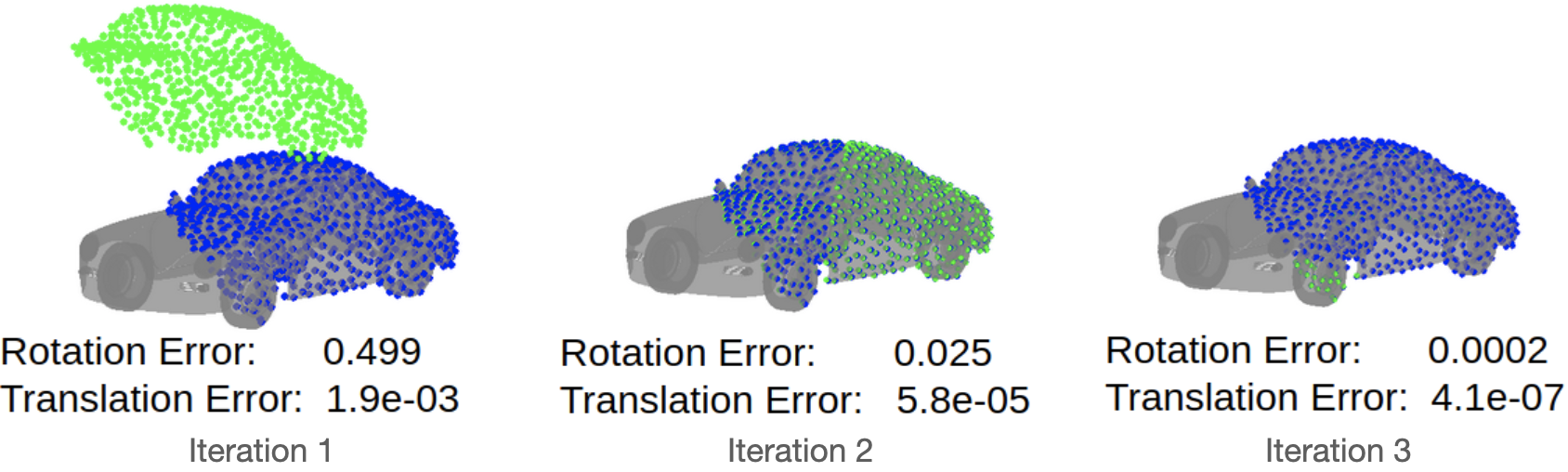}
    \end{center}
   \caption{Qualitative results for Section \ref{sec:mask_evaluation}. For each iteration, we first use MaskNet to compute mask and then PointNetLK to compute registration parameters. The source point cloud is shown in green color, template in gray color and point cloud registered with PointNetLK in blue color.  With every iteration, we observe a continuous decrease in rotation as well as translation error.}
\label{fig:iteration_results}
\end{figure}

Prior works~\cite{yaoki2019pointnetlk, vsarode2019pcrnet, yuan2018iterative} show the sensitivity of PointNet to large initial misalignment between a given pair of point clouds. Taking this into consideration, we evaluate the mask in an iterative manner where we first compute mask $\mathcal{C}$ and then estimate $\mathbf{R}$ and $\mathbf{t}$ by substituting $\mathcal{C}$ in Eq. \ref{eq:feature_compare}. We train MaskNet to estimate mask ($\mathcal{C}$) and separately train PointNetLK to estimate registration parameters between a pair of point clouds from ModelNet40 dataset using all 40 object categories. Fig.~\ref{fig:iteration_results} shows the results of estimated registration parameters for each iteration. We observe a monotonic decrease in rotation as well as translation error in each iteration. Even though $\mathcal{C}, \mathbf{R},$ and $\mathbf{t}$ improve with every iteration, we observe results better than the state-of-art partial registration methods in the first iteration with minimum computational effort. For that reason, in all the further experiments we only report results for a single iteration.

\begin{figure}[htbp]
    \begin{center}
        \includegraphics[width=0.8\linewidth]{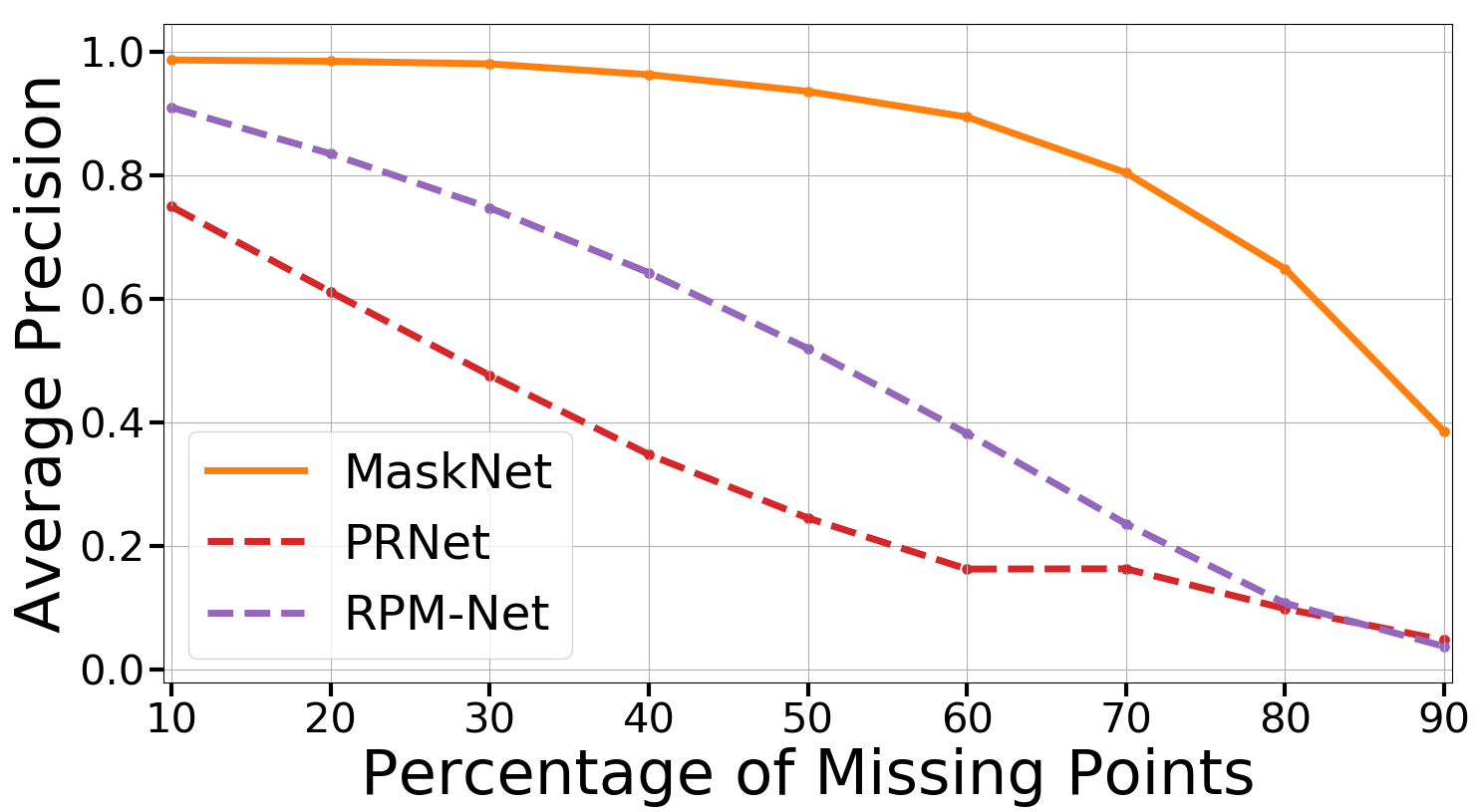}
    \end{center}
   \caption{Results for section \ref{sec:mask_evaluation}. The plot shows the precision of predicted inliers on the y-axis and percentage of missing points in the partial point cloud on x-axis. Mask produced by MaskNet and correspondences produced by PRNet, RPM-Net are used to compute inliers. Even though MaskNet is trained with 30 percent missing points, we observe a reasonably good performance in the range of 10 to 60\% missing points as compared to other algorithms.}
\label{fig:precision_vs_missing_pts}
\end{figure}

\begin{figure}[h]
    \centering
    \begin{subfigure}{.165\textwidth}
        \centering
        \includegraphics[width=\linewidth]{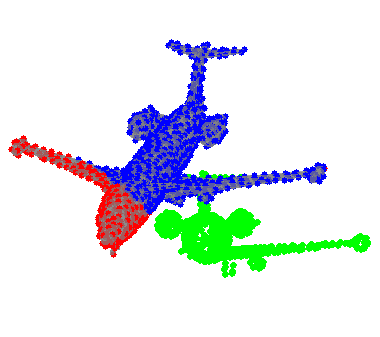}
        \caption{Seen Category}
        \label{fig:mask_qualitative_results_a}
    \end{subfigure}
    \begin{subfigure}{.15\textwidth}
        \centering
        \includegraphics[width=\linewidth]{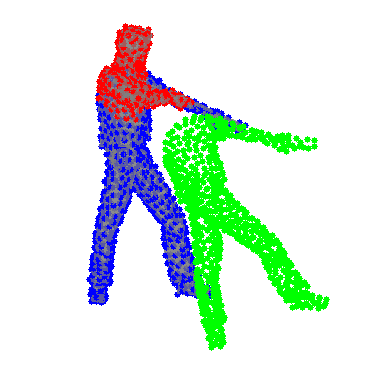}  
        \caption{Unseen Category}
        \label{fig:mask_qualitative_results_c}
    \end{subfigure}
    \begin{subfigure}{.15\textwidth}
        \centering
        \includegraphics[width=\linewidth]{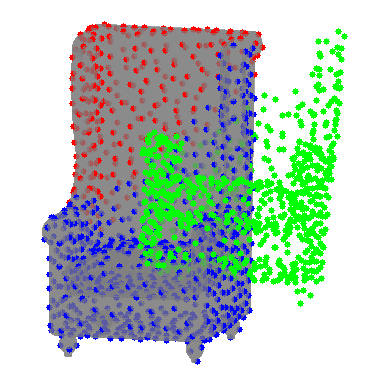}  
        \caption{Unseen Category}
        \label{fig:mask_qualitative_results_d}
    \end{subfigure}
    \caption{Results showing the template point cloud $\mathcal{X}$ visualized as a CAD model, and source point cloud $\mathcal{Y}$ as green colored points. Blue point cloud represents $\mathcal{X}_1=\mathcal{C} \otimes \mathcal{X}$ and points from $\mathcal{X}$ having $\mathcal{C}_i=0$ are shown in red color.}
    \label{fig:mask_qualitative_results}
\end{figure}

MaskNet computes a mask giving the probabilities for each point in the full point cloud for having a similar point in the partial ones. Since each element in the mask can be either $0$ or $1$, it becomes a binary classification problem for each point. We use the "precision metric" which is a common metric in binary classification, to quantify the efficiency of our method. 
Inlier points having $\mathcal{C}^{gt}_i=1$ are termed as true positives (TP) and those having $\mathcal{C}^{gt}_i=0$
are termed as false positives (FP). 
\begin{equation}
    Precision = \frac{TP}{TP + FP}
\end{equation}
In this experiment, we train MaskNet, PRNet and RPM-Net on the entire ModelNet40 dataset including all 40 object categories with $30\%$ missing points in partial point clouds. Ground truth mask and the prediction of MaskNet is used to estimate TP and FP. PRNet and RPM-Net provide correspondence matrix as output which are used to find inlier point pairs (TP) and outliers.
To evaluate our method, we compute precision using a test set with different percentages of missing points in input point clouds as shown in Fig.~\ref{fig:precision_vs_missing_pts}. We observe that MaskNet performs reasonably well within a range of $10\%$ to $60\%$ of missing data when compared to PRNet and RPM-Net. The decreasing precision of PRNet and RPMNet (as opposed to MaskNet) is attributed to the sensitiveness of Horn's method that is used to find pose parameters from incorrect correspondences. \\
Fig.~\ref{fig:mask_qualitative_results} shows the inlier points (blue colored) computed using the predicted mask for the given partial point cloud (green colored) and a full point cloud (gray CAD model). The results clearly indicate the accuracy of the mask from the similarity in the geometric shape of blue and green point clouds.

\subsection{Point Cloud Denoising}
\label{sec:denoising}
In this section, we show the use of MaskNet to detect the outliers in a given point cloud by comparing it with a standard reference point cloud of the same object category. We train MaskNet with one of the input point clouds from the ModelNet40 dataset having $10\%$ points added as outliers at random locations in 3D space. While training, the ground truth mask $\mathcal{C}^{gt}$ contains $0$ for all indices that are outlier points and $1$ for inliers.
Our training dataset consists of all point clouds from the training set of ModelNet40 while  our testing dataset has 1000 randomly chosen point clouds from the test set.

\begin{figure}[h!]
    \begin{center}
        \includegraphics[width=0.8\linewidth]{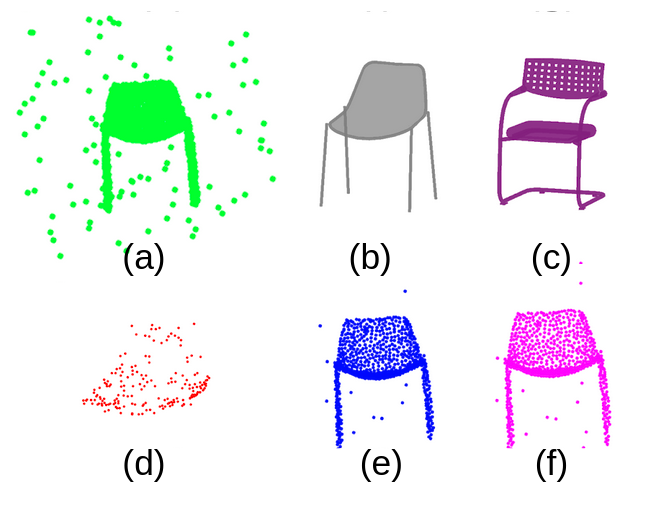}
    \end{center}
    \caption{Denoising with different templates models. (a) Noisy source, (b) and (c) are different models of category 'chair'. (d) Result of PointCleanNet~\cite{rakotosaona2020pointcleannet}, (e) and (f) are results of MaskNet when using template models (b) and (c) respectively with source (a).}
\label{fig:outlier qualitative results.}
\end{figure}

\begin{figure}[h!]
    \centering
    \begin{subfigure}{.23\textwidth}
        \centering
        \includegraphics[width=\linewidth]{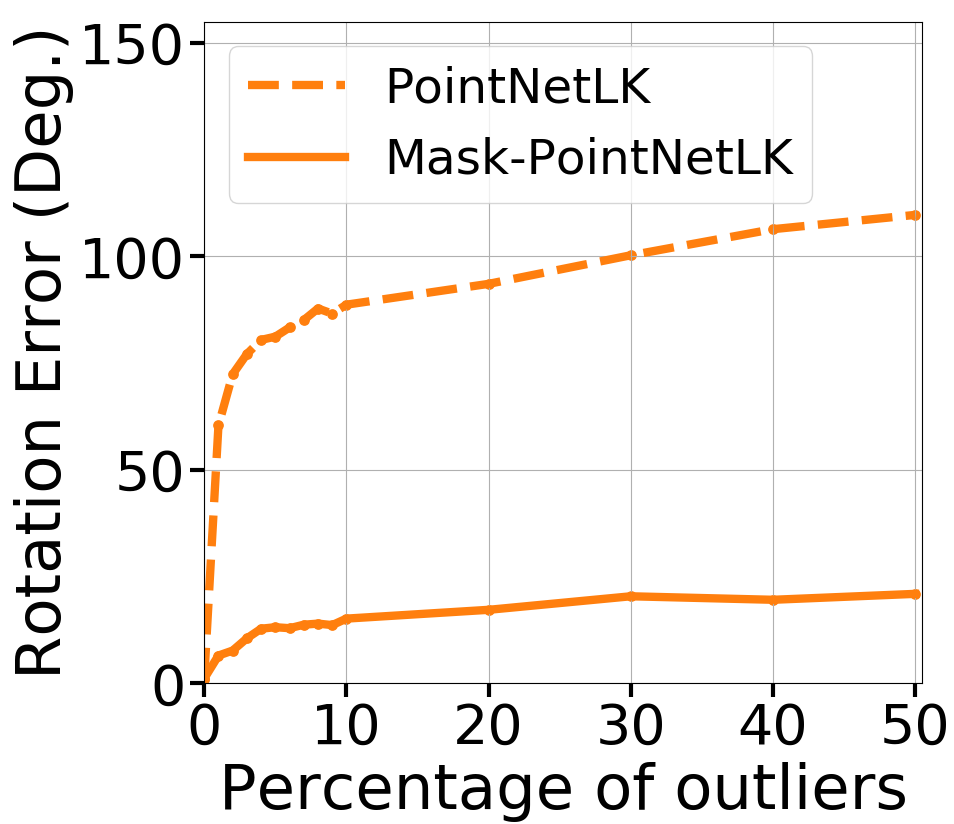}  
        \caption{}
        \label{fig:sub-first}
    \end{subfigure}
    \begin{subfigure}{.23\textwidth}
        \centering
        \includegraphics[width=\linewidth]{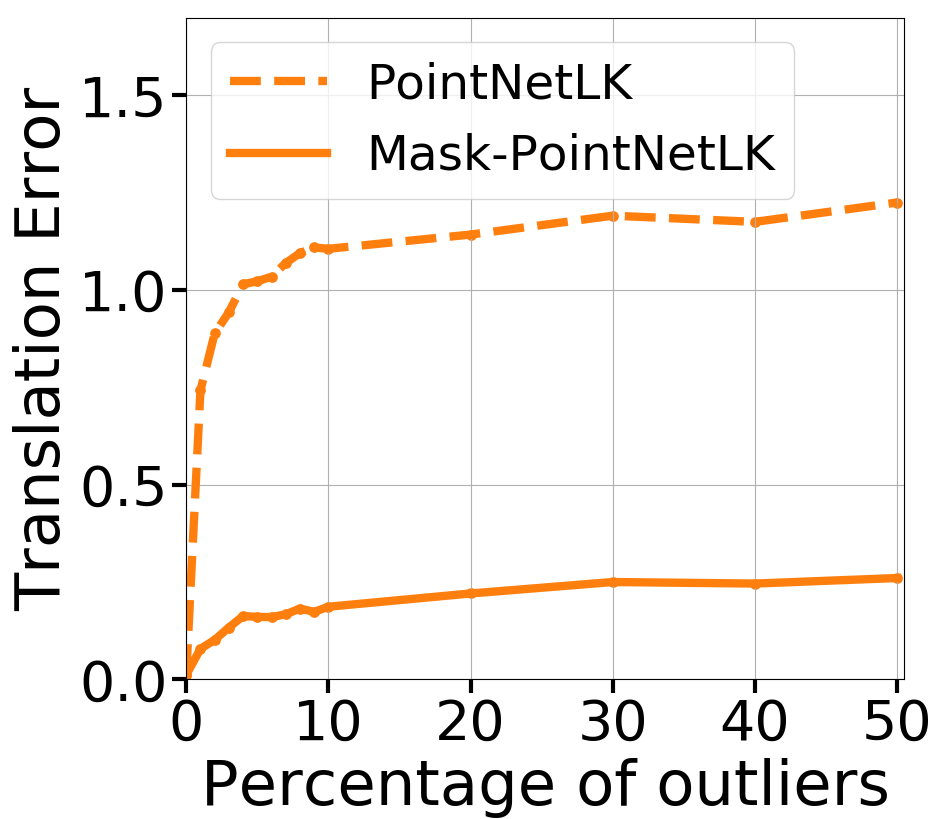}  
        \caption{}
        \label{fig:sub-first}
    \end{subfigure}
    \caption{Results for Section~\ref{sec:denoising} showing the effect of number of outliers for PointNetLK and its retrofitted version with MaskNet using rotation and translation errors as a metric for comparison. MaskNet is trained with only 10\% outliers but shows a great improvement in registration performance of Mask-PointNetLK.}
    \label{fig:outlier_registration}
\end{figure}

Fig.~\ref{fig:outlier qualitative results.}~(e,f) shows MaskNet's ability to use different template models (of the same category as that of the noisy source point cloud) to denoise the source point cloud. Fig.~\ref{fig:outlier qualitative results.}~(a,d) also shows the performance of PointCleanNet~\cite{rakotosaona2020pointcleannet}  on the same noisy source point cloud. While PointCleanNet removes most outlier points, it also removes a fair bit of inliers.
In Fig.~\ref{fig:outlier_registration}, we show that retrofitting PointNetLK with MaskNet improves its performance significantly. Even 3\% to 4\% percentage of outliers increases the rotation error by $\approx 70^\circ$ for PointNetLK compared to PointNetLK when retrofitted with MaskNet (referred to as Mask-PointNetLK). A similar trend is observed for translation error. This clearly shows the advantage of denoising the point cloud before registration using our network.

\subsection{Partial Point Cloud Registration}
\label{sec:registration}
\begin{figure*}
    \centering
    \begin{subfigure}{.23\textwidth}
        \centering
        \includegraphics[width=\linewidth]{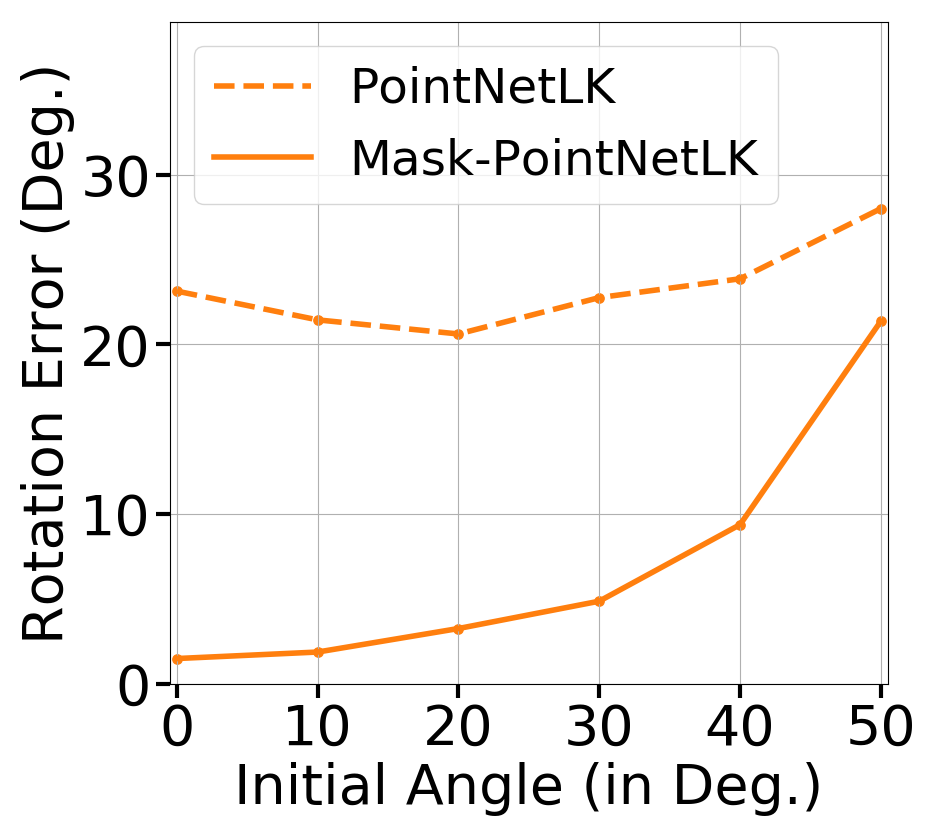}  
        \caption{}
        \label{fig:sub-first}
    \end{subfigure}
    \begin{subfigure}{.23\textwidth}
        \centering
        \includegraphics[width=\linewidth]{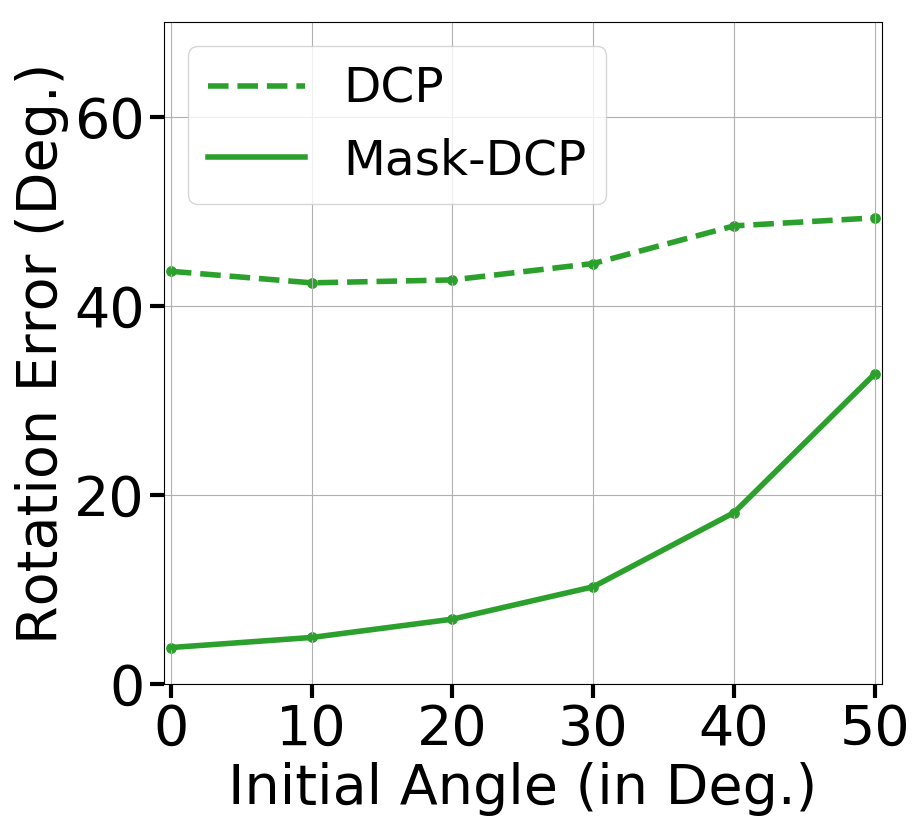}  
        \caption{}
        \label{fig:sub-first}
    \end{subfigure}
    \begin{subfigure}{.23\textwidth}
        \centering
        \includegraphics[width=\linewidth]{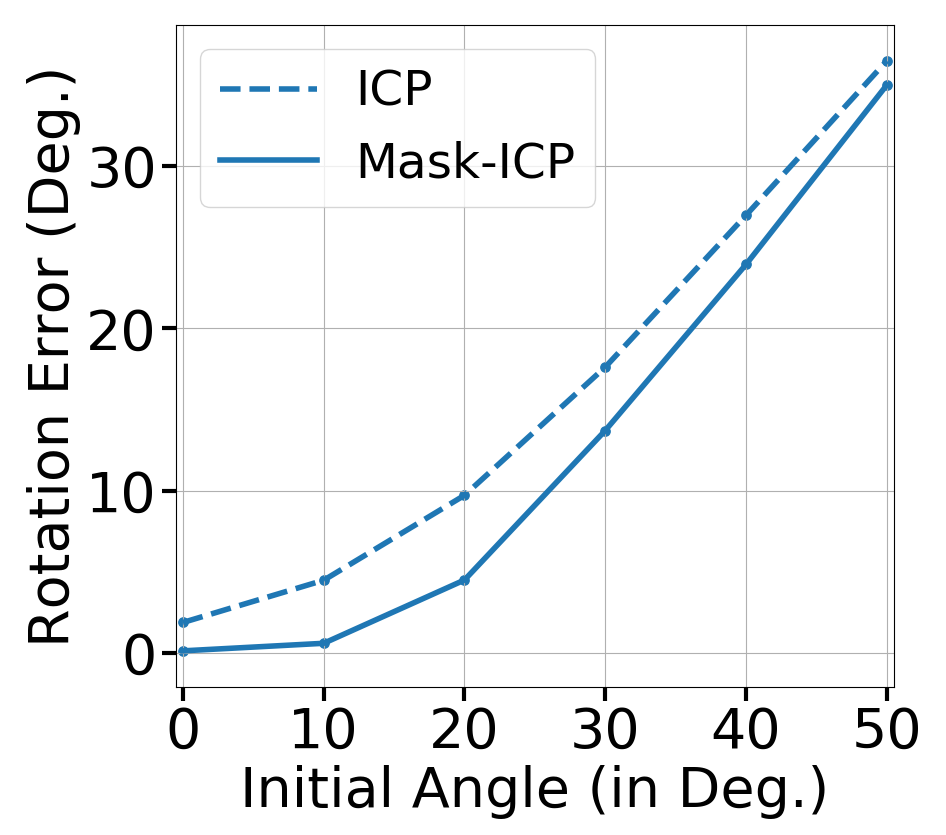}  
        \caption{}
        \label{fig:sub-first}
    \end{subfigure}
    \begin{subfigure}{.23\textwidth}
        \centering
        \includegraphics[width=\linewidth]{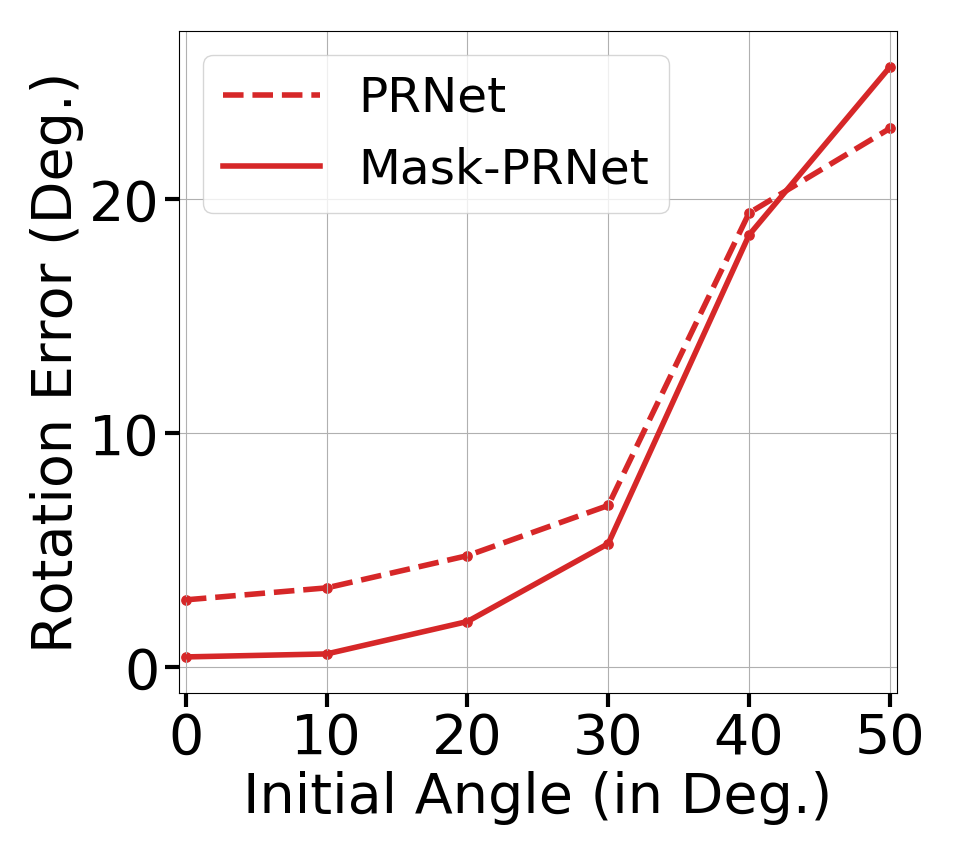}  
        \caption{}
        \label{fig:sub-first}
    \end{subfigure}

    \begin{subfigure}{.23\textwidth}
        \centering
        \includegraphics[width=\linewidth]{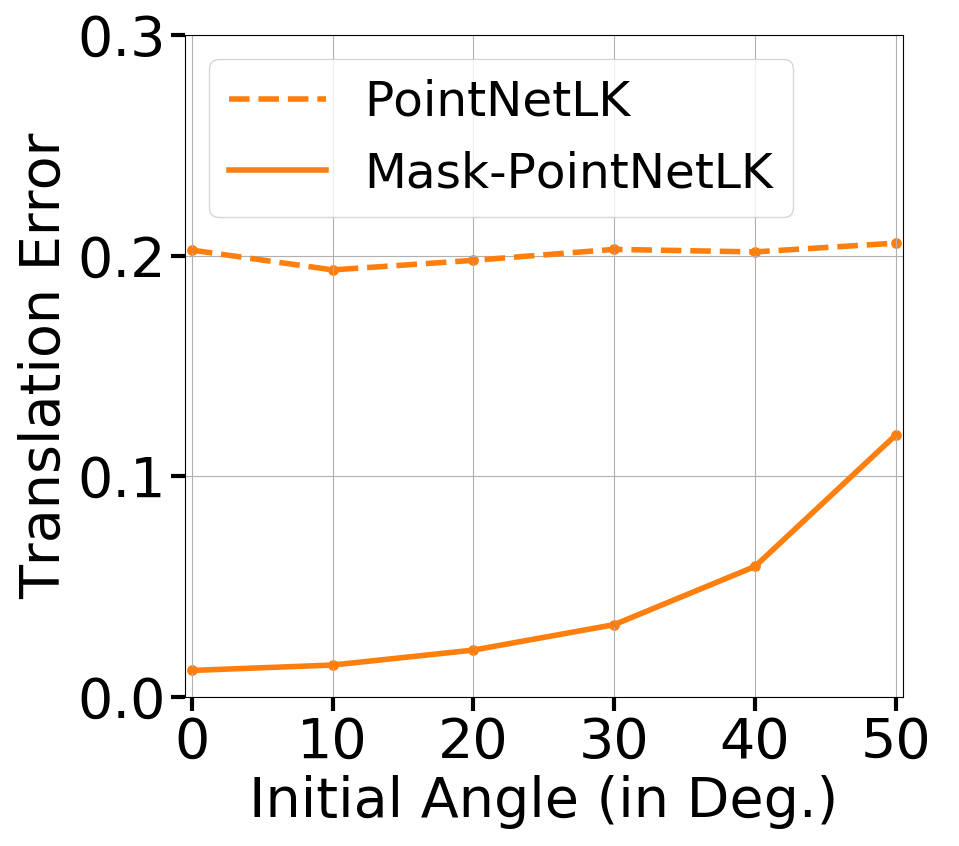}  
        \caption{\\PointNetLK}
        \label{fig:sub-first}
    \end{subfigure}
    \begin{subfigure}{.23\textwidth}
        \centering
        \includegraphics[width=\linewidth]{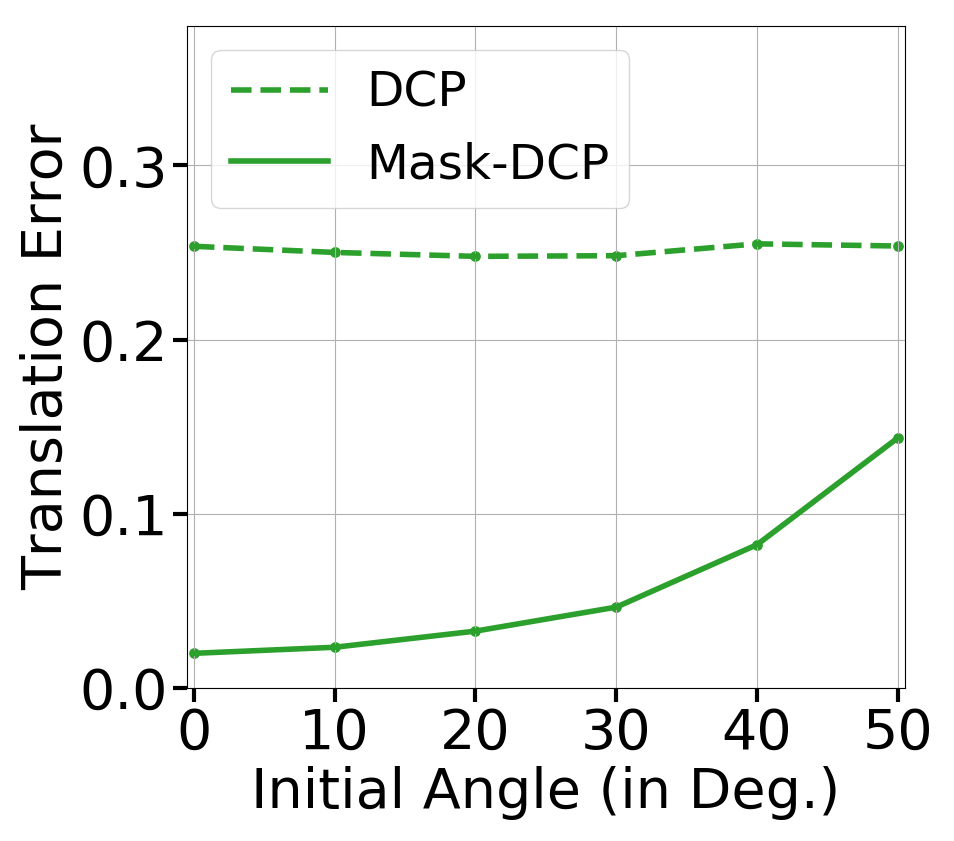}  
        \caption{\\DCP}
        \label{fig:sub-first}
    \end{subfigure}
    \begin{subfigure}{.23\textwidth}
        \centering
        \includegraphics[width=\linewidth]{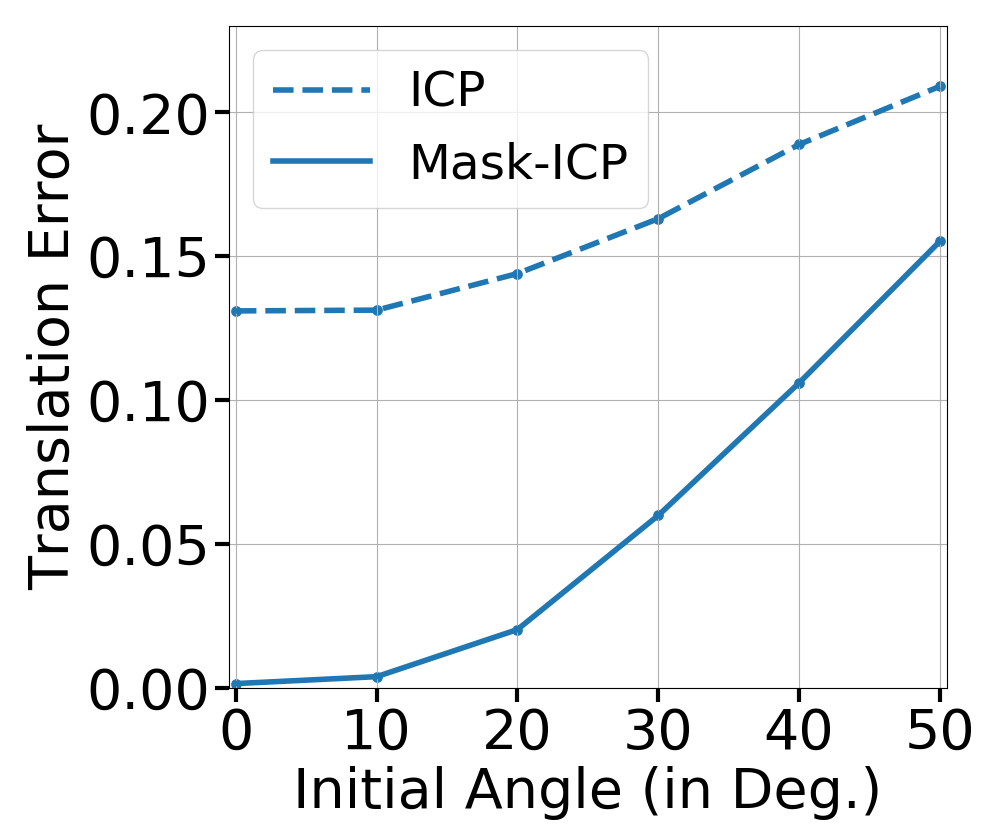}  
        \caption{\\ICP}
        \label{fig:sub-first}
    \end{subfigure}
    \begin{subfigure}{.23\textwidth}
        \centering
        \includegraphics[width=\linewidth]{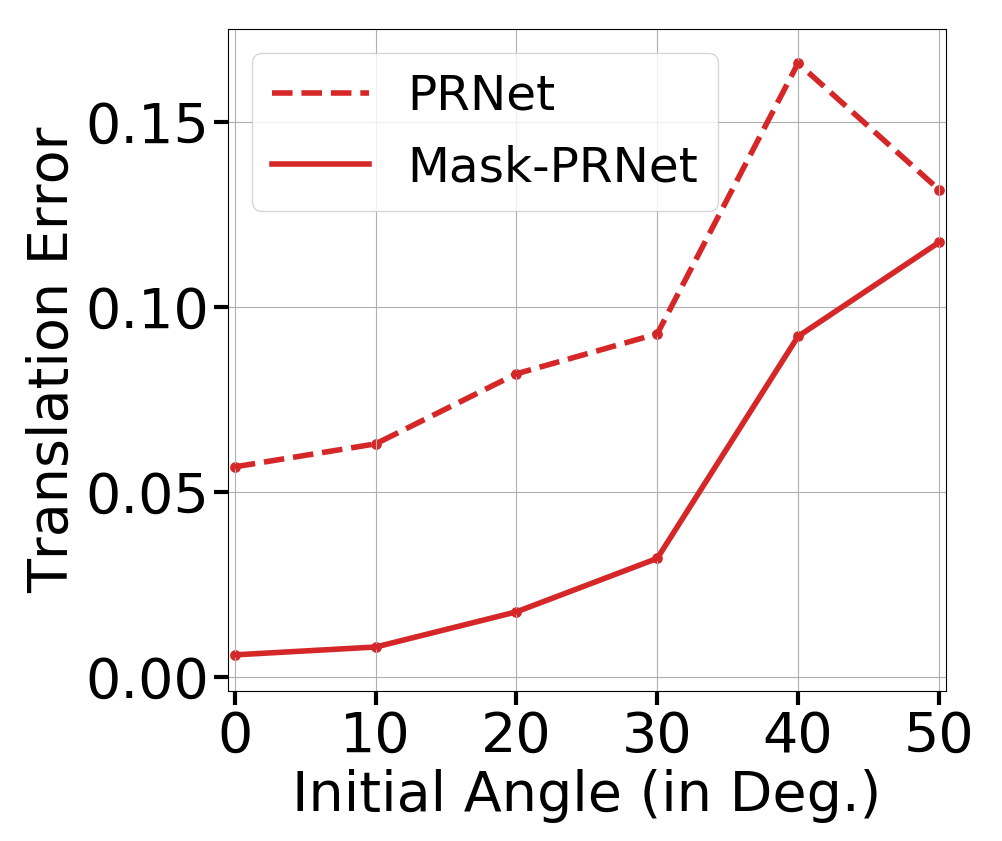}  
        \caption{\\PRNet}
        \label{fig:sub-first}
    \end{subfigure}
    \caption{Results for section \ref{sec:registration}. Effect of initial misalignment on rotation error (top row) and translation error (bottom row) for X and Mask-X [X: PointNetLK, DCP, ICP].
    Figures (a) and (b) show an improvement of approximately $25^\circ$ in rotation error due to the use of MaskNet with PointNetLK and DCP. Similarly, improvement in translation errors can be observed in (e) and (f). Figures (c) and (g) show that Mask-ICP has a marginal improvement in registration over ICP. Similarly, figures (d) and (h) show improvement in registration using Mask-PRNet over PRNet.}
    \label{fig:partial point cloud registration}
\end{figure*}

We use partial point cloud registration as a downstream task to show the usefulness of MaskNet. The registration pipeline in this experiment is as shown in Fig.~\ref{fig:masknet_registration_framework}. The pipeline consists of two stages -- (1) the MaskNet estimates the inlier points from given input point clouds, and (2) a standard registration algorithm (classical or deep learning-based) estimates the registration parameters. Training of neural networks in both the stages takes place independently. We train MaskNet with the entire ModelNet40 dataset and conduct experiments using ICP~\cite{INTRO:ICP}, PointNetLK~\cite{yaoki2019pointnetlk}, DCP~\cite{Wang_2019_ICCV} and PRNet~\cite{Wang_2019_NeurIPS}. Pre-trained models of PointNetLK, PRNet and DCP are used in this pipeline. As ICP and PointNetLK are iterative methods, we allow maximum of 10 iterations for all the following experiments.

\begin{figure}[h!]
    \centering
    \includegraphics[width=0.75\linewidth]{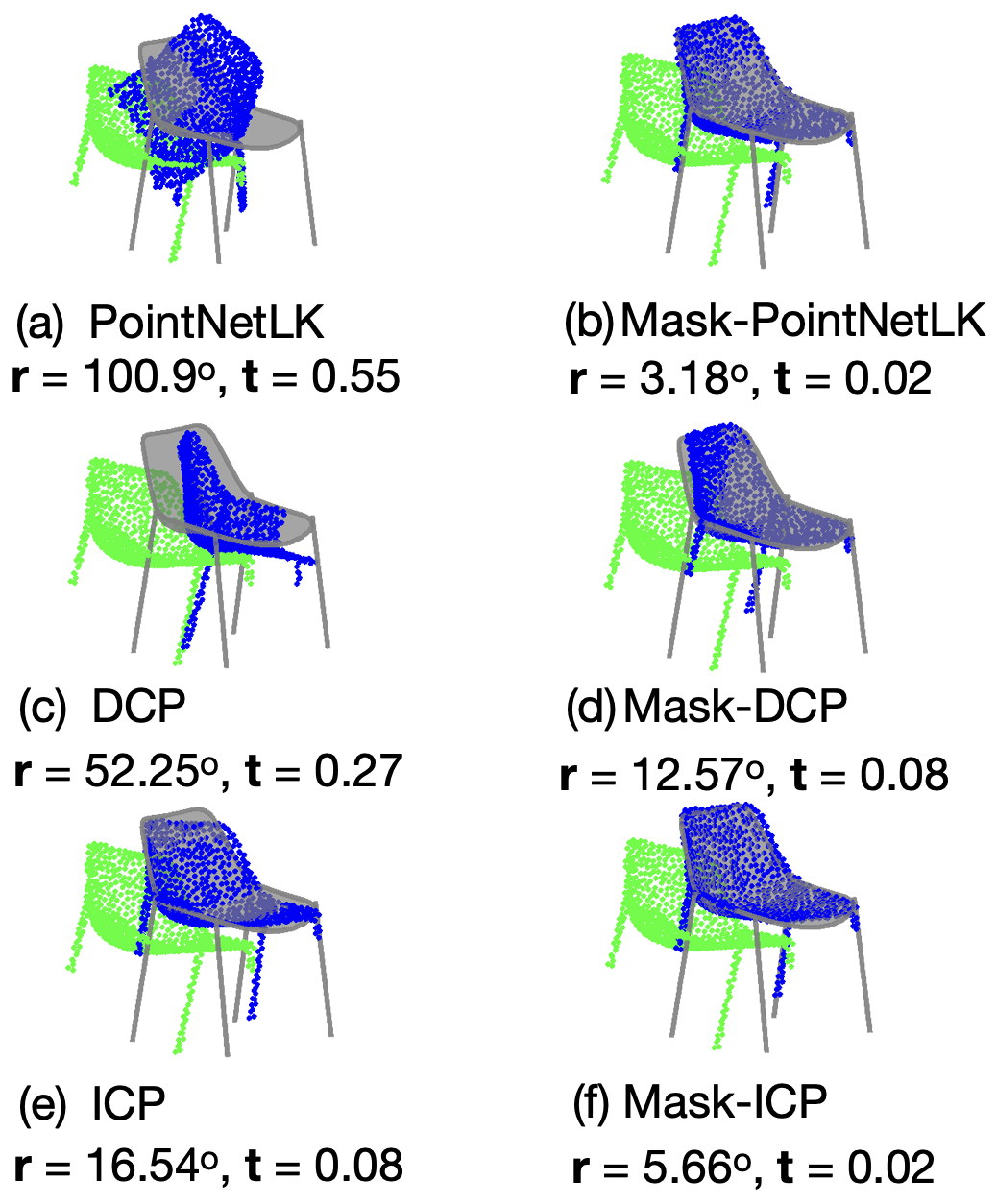}  
    \caption{Improvement in registration through MaskNet. The left column shows registration by 3 methods and the right column shows same three methods when augmented by MaskNet. \textbf{r}: Rotation Error, \textbf{t}: Translation Error}
     \label{fig:mask_quantitative_results}
\end{figure}

In this experiment, we compare ICP, PointNetLK, PRNet and DCP algorithms with their MaskNet equipped versions hereafter referred to as Mask-ICP, Mask-PointNetLK, Mask-PRNet and Mask-DCP, respectively. Experimental settings to create evaluation dataset are same as described in Sec.~\ref{sec:mask_evaluation} and the dataset consists of 1000 pairs of point clouds. Mean rotation and translation errors for different initial misalignments are reported in the Fig.~\ref{fig:partial point cloud registration}. PointNetLK and DCP perform poorly with the partial point clouds. We see a notable performance improvement in the error metrics of Mask-PointNetLK and Mask-DCP as compared to PointNetLK and DCP respectively. As ICP is a shape agnostic classical registration method, Mask-ICP shows a marginal performance improvement over ICP. Even though PRNet is specifically designed for partial point cloud registration, we observe an improvement in the performance when using MaskNet. Qualitative results are shown in Fig.~\ref{fig:mask_quantitative_results}. Average computational time for ICP, PointNetLK, PRNet and DCP is 6.82 ms, 52.30 ms, 68.52 ms and 23.99 ms. On the other hand, Mask-ICP, Mask-PointNetLK, Mask-PRNet and Mask-DCP requires 9.83 ms, 58.15 ms, 82.18 ms and 24.68 ms showing a negligible increase in time complexity with the addition of MaskNet.

\begin{figure}[h!]
    \centering
    \includegraphics[width=0.95\linewidth]{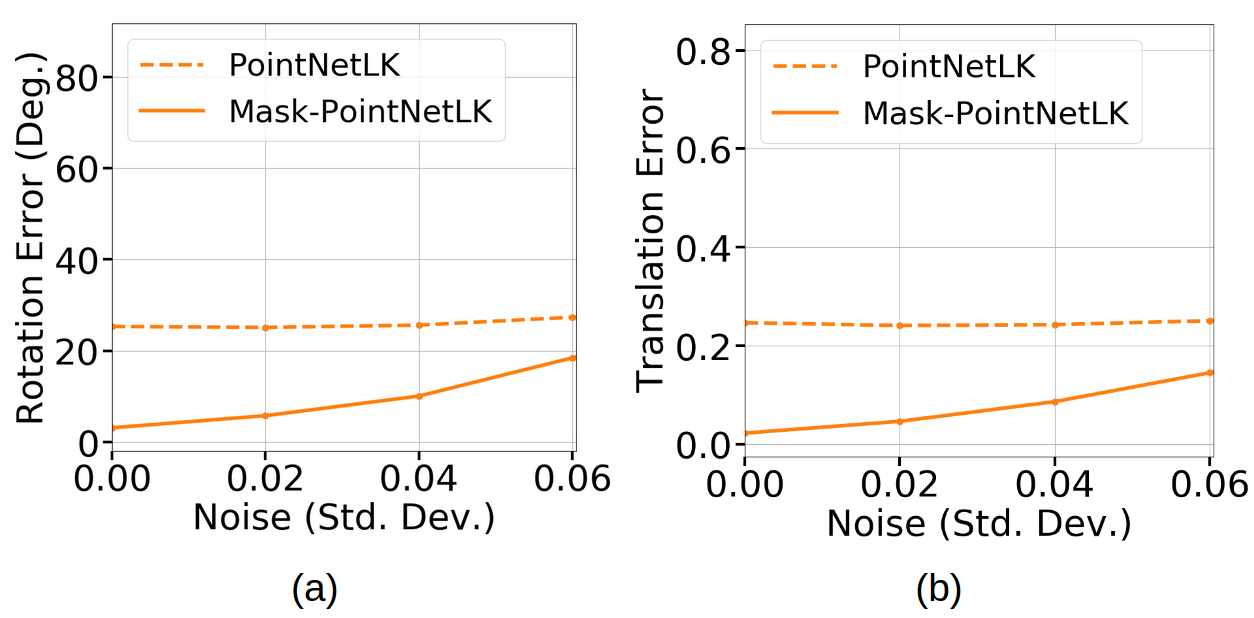}  
    \caption{Robustness to gaussian noise during registration. MaskNet affords robustness to presence of gaussian noise, as is seen by the improvement of registration metrics ((a) mean rotation error and (b) mean translation error) brought on by using Mask-PointNetLK over PointNetLK.}
     \label{fig:noise}
\end{figure}

Further, MaskNet is robust to the presence of gaussian noise during registration. This is demonstrated (in Fig.~\ref{fig:noise}) through a comparison of registration results from PointNetLK and Mask-PointNetLK on partial point clouds containing gaussian noise. Each point of a partial point cloud (from the ModelNet40 dataset) is perturbed with noise sampled from a gaussian of zero-mean and standard deviation in the range of 0 to 0.06. PointNetLK and MaskNet are separately trained on the perturbed ModelNet40 dataset and evaluation performed on 1000 pairs of point clouds from the 40 object categories of ModelNet40 and the perturbed partial point clouds. Fig.~\ref{fig:noise} shows that Mask-PointNetLK has lower rotation as well as translation error as compared to PointNetLK.

In conclusion, MaskNet proves to be an efficient method to deal with partial point cloud registration when used with any existing registration algorithm. An interesting property of this pipeline is that both the MaskNet and the registration algorithm are independent of each other and can be trained separately. This reduces the computational efforts required to retrain any registration network.

\subsection{Generalization}
\label{sec:generalization}
\begin{figure}[ht]
    \centering
    \begin{subfigure}{.21\textwidth}
        \centering
        \includegraphics[width=\linewidth]{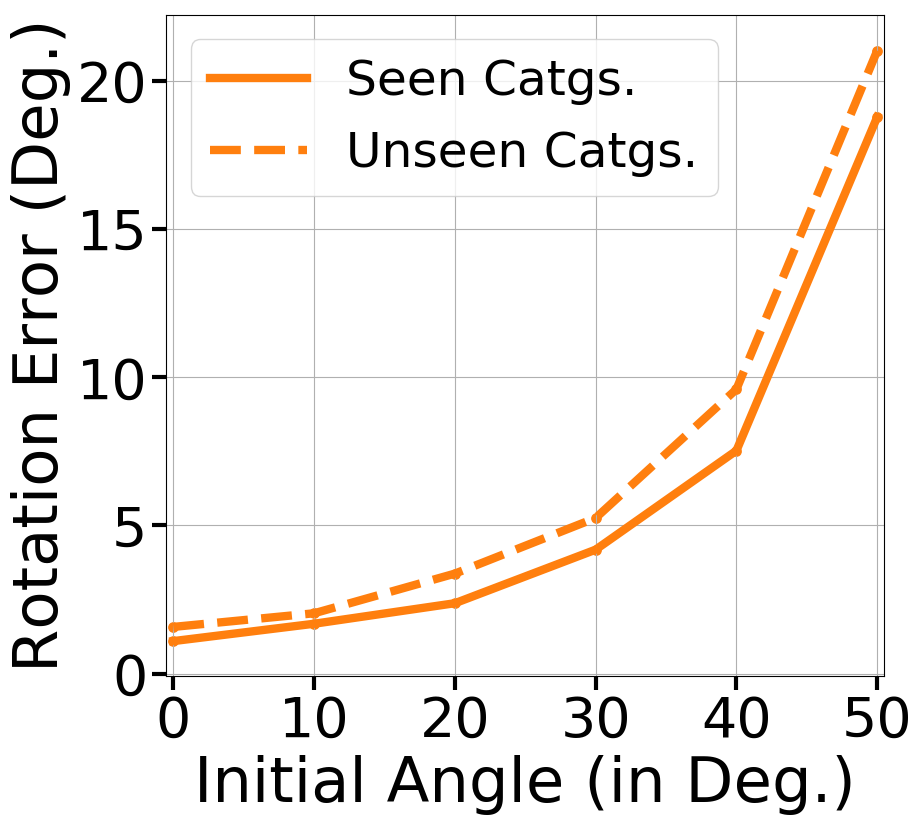}  
        \caption{}
        \label{fig:sub-first}
    \end{subfigure}
    \begin{subfigure}{.23\textwidth}
        \centering
        \includegraphics[width=\linewidth]{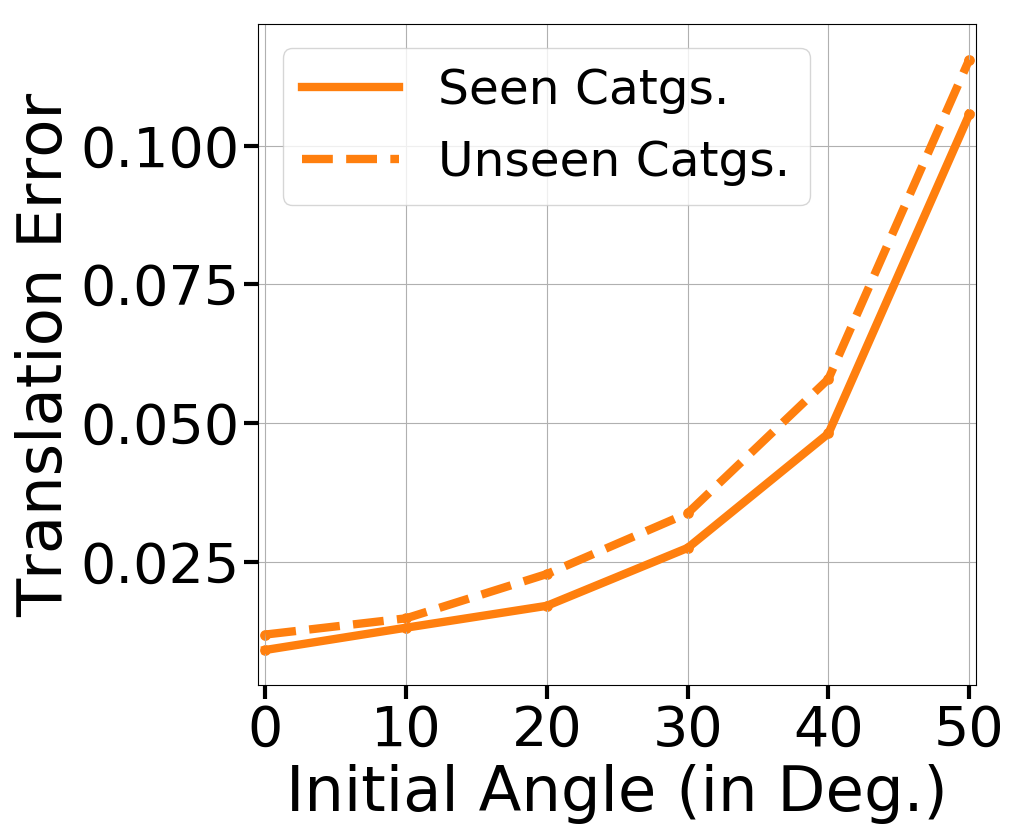}  
        \caption{}
        \label{fig:sub-first}
    \end{subfigure}
    
    \caption{Results showing the generalization of MaskNet by training the network with 20 object categories (Seen) from ModelNet40 dataset and testing with 20 different object categories (Unseen). The above plots show that there is a minimal difference in the performance of Mask-PointNetLK for seen and unseen categories. This clearly indicates the generalization capability of MaskNet across unseen object categories.}
    \label{fig:registration generalization}
\end{figure}

We split ModelNet40 dataset into two parts -- models of first 20 categories for training (seen categories) and the last 20 categories for evaluation (unseen categories). Point cloud datasets are created for both the categories using the protocol described in Sec.~\ref{sec:mask_evaluation}. MaskNet is trained only using the seen categories and evaluated using the point clouds from unseen categories. Qualitative results of evaluated mask on unseen categories is shown in Fig.~\ref{fig:mask_qualitative_results}, where blue points indicate the estimated inlier points and red points show the outliers. We observe that the geometric shape of green and blue point clouds are in good accordance, which indicates the prediction of an accurate mask even in case of unseen categories.
We further evaluate MaskNet by estimating the registration between partial point clouds. We choose a test-set of 1000 point clouds each from seen as well as unseen categories and perturb them with various initial misalignment. We then compute the mean rotation error and mean translation error for both the datasets after registering them with Mask-PointNetLK. The small difference in the performance of Mask-PointNetLK for seen versus the unseen categories in Fig.~\ref{fig:registration generalization}, clearly indicates the generalization ability of MaskNet across different object categories.

\subsection{Real-world Data}
\label{sec:real_world_experiments}
\begin{figure}[t]
    \begin{subfigure}{.45\textwidth}
        \centering
        \includegraphics[width=\linewidth]{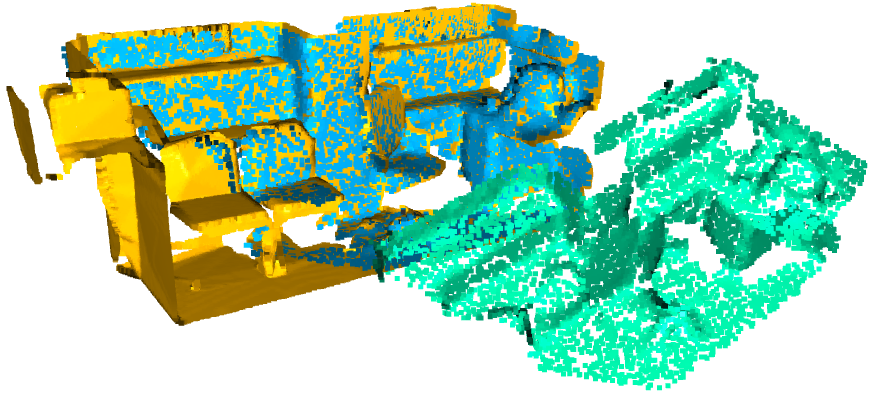}
        \caption{S3DIS Result}
        \label{fig:sub-first}
    \end{subfigure}
    \begin{subfigure}{.45\textwidth}
        \centering
        \includegraphics[width=\linewidth]{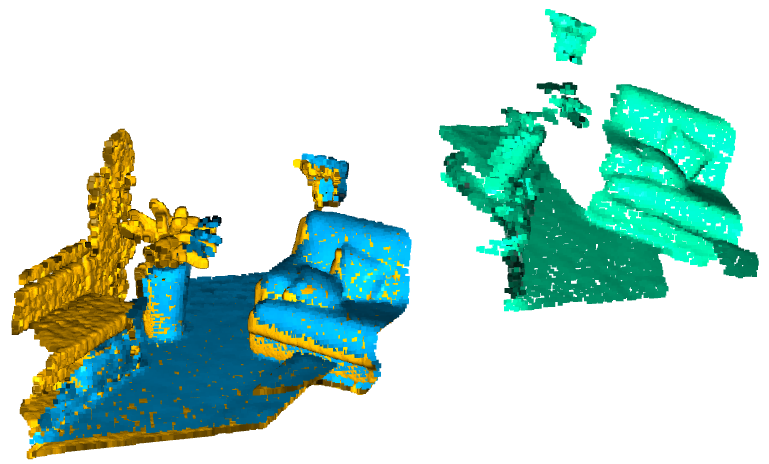}
        \caption{3DMatch Result}
        \label{fig:sub-first}
    \end{subfigure}
    \caption{Qualitative results for real world 3D scans. Green is the partial point cloud ($\mathcal{Y}$), yellow is the CAD model from which a full point cloud is sampled uniformly ($\mathcal{X}$), and blue point cloud is the result of Mask-PointNetLK registration pipeline.}
    \label{fig:real world results.}
\end{figure}

\begin{figure}[h]
    \centering
    \includegraphics[width=0.75\columnwidth]{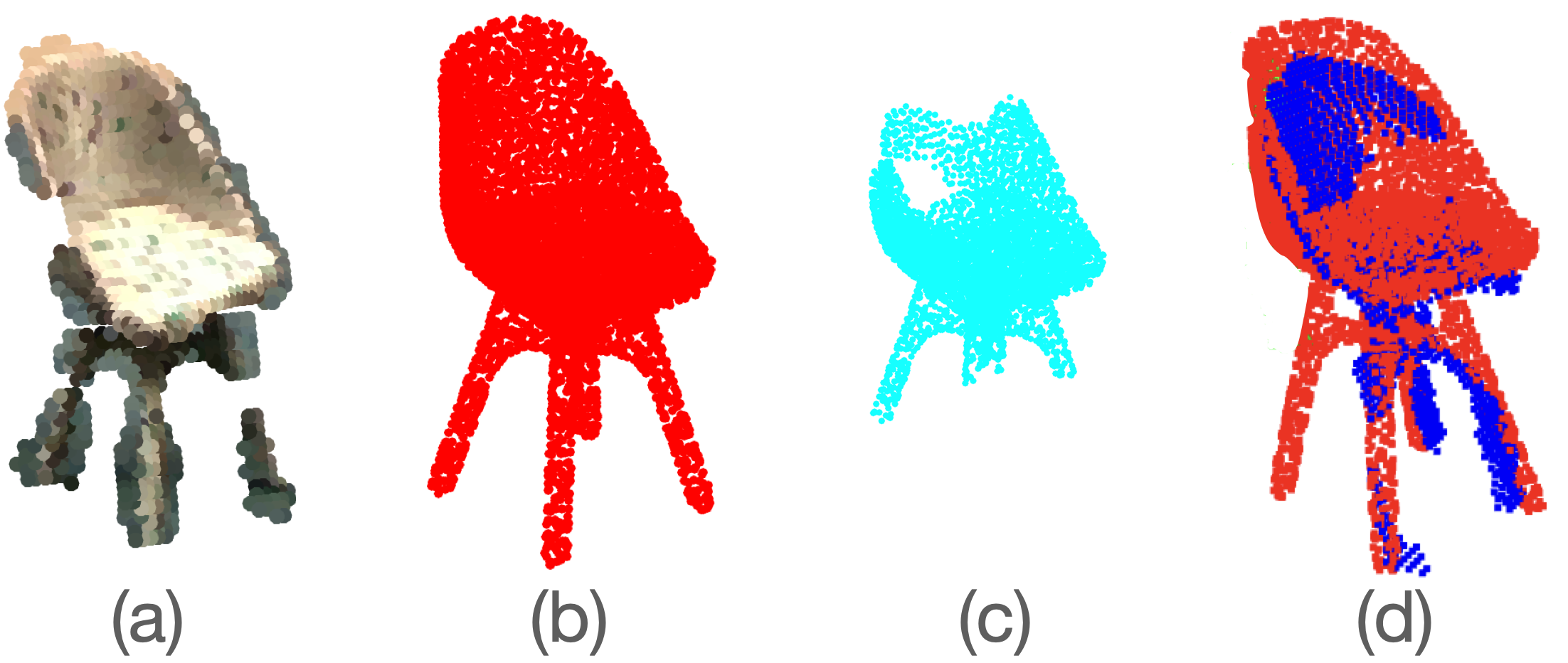}
    \caption{(a) Point cloud obtained from a RealSense sensor. (b) Template point cloud of the chair (red), (c) Result of MaskNet (cyan), (d) blue point cloud is the result of Mask-PointNetLK.}
    \label{fig:realworld}
\end{figure}

MaskNet shows good performance for zero shot transfer of learning across the S3DIS~\cite{armeni_cvpr16} and 3DMatch datasets~\cite{zeng20163dmatch}. S3DIS contains a single large point cloud consisting of various offices, conference rooms, kitchen, etc. We divided this large point cloud into smaller point clouds and created a processed-S3DIS dataset. We use this processed-S3DIS dataset to train MaskNet, and tested using point clouds present in the 3DMatch dataset. We observe that MaskNet trained on S3DIS performs equally well on 3DMatch dataset without the need of any additional fine-tuning. This can be clearly observed in Fig.~\ref{fig:real world results.}, where yellow and green point clouds are the input to Mask-PointNetLK and blue point cloud shows the registered point cloud.MaskNet was also tested on point clouds of 3D printed objects, obtained from a RealSense sensor. Qualitative results are shown in Fig.~\ref{fig:realworld}.

\section{Conclusion \& Future Work}
We offer a new learning-based approach, MaskNet, for determining inlier points in a given pair of point clouds by computing a \textit{mask}. Our approach has a higher accuracy of finding inliers compared to existing learning-based inlier estimation methods which are also computationally expensive. We demonstrate through experiments, that MaskNet -- \textit{(a)} augments the ability of existing classical and deep learning-based registration methods to better deal with partial point clouds and outliers, \textit{(b)} can be used to reject noise, and \textit{(c)} generalizes to object categories that it was not trained on.

While we currently use a PointNet encoding in MaskNet, in the future we could replace PointNet with other feature descriptors that are less sensitive to noise, and are invariant to pose transformations. A natural next step, from the perspective of real-world application, is to remove supervision on ground truth masks - as unsupervised networks tend to generalize far better than supervised networks. In addition, this would allow us to train directly on real world datasets without the need to hand label the inlier points. 

MaskNet is currently limited to removing points from only one of the input point clouds. A logical extension would be to estimate two sets of inlier points - one for each point cloud - in a given point cloud pair. This can be helpful in tasks involving stitching point clouds for scene reconstruction and SLAM applications. MaskNet offers a starting point for further development of such tasks and applications.

{\small
\bibliographystyle{ieee}
\bibliography{egbib}
}

\end{document}